\documentclass[11pt,letter]{article}

\usepackage{poma_style}

\usepackage{graphicx}
\usepackage{subcaption}
\usepackage{multirow}
\usepackage{amsmath,amsfonts,bm}
\usepackage{courier}
\usepackage[colorlinks=true,urlcolor=blue,linkcolor=black,citecolor=black]{hyperref}

\title{Identifying the nonlinear string dynamics with \\port-Hamiltonian neural networks}
\author{
Maximino Linares$^{1,2}$, Guillaume Doras$^{1}$, Thomas Hélie$^{1,3}$ \\[0.5em]
{\small $^{1}$IRCAM, 1, place Igor-Stravinsky, 75004, Paris, France}\\
{\small $^{2}$Sorbonne Université, 4, place Jussieu, 75005, Paris, France}\\
{\small $^{3}$CNRS, 3 rue Michel-Ange, 75016, Paris, France}
}

\begin{document}

\maketitle

\begin{abstract}
    Hybrid machine learning combines physical knowledge with data-driven models to enhance interpretability and performance. In this context, Port-Hamiltonian Systems (PHS), which generalize Hamiltonian mechanics to describe open, non-autonomous dynamical systems, have been successfully integrated with neural networks under the name Port-Hamiltonian Neural Networks (PHNNs). While the ability of PHNNs to identify Hamiltonian ordinary differential equation (ODE) systems has already been demonstrated, their application to learning Hamiltonian partial differential equation (PDE) systems remains largely unexplored. This limitation restricts their use in musical acoustics, where instruments are typically modeled as distributed parameter systems governed by PDEs. In this work, we demonstrate how to learn the nonlinear string dynamics from data in a physically-consistent framework through a PHNN extension to PDEs. By constructing structured neural network architectures based on PHS, we can recover both the Hamiltonian governing the string and the dissipation affecting it. This approach outperforms baseline, non-physics-informed methods in terms of both accuracy and interpretability. Numerical experiments using synthetic data demonstrate the ability of the proposed PHNN model to identify and emulate the nonlinear dynamics of the system.
\end{abstract}

\section{Introduction}
Estimation, identification, and learning all describe similar notions, even though different disciplines use different terms~\cite{carlson2019machine}. In machine learning, the goal is to learn models from data that can predict the output given an unseen input~\cite{murphy2012machine}. In control theory, system identification experiments with a system and uses its responses to build a model that agrees with the observations~\cite{ljung1999system}. Both fields share the same objective: creating models that capture the underlying patterns in data. Determine which model to use is complex and is normally driven by the specific goals of the application. Linear models are especially useful for systems operating near an operating point, as the behavior of a nonlinear system can often be approximated locally by a linear model~\cite{glad2018control}. When linear models fail, gray-box models use known structure to fit the unknown parameters to data~\cite{johansson1993system}. They require domain knowledge and can be hard to build, but offer deeper understanding when successful. On the other hand, black-box models are flexible and can fit almost any data~\cite{hornik1989multilayer}. However, they lack interpretability and usually risk overfitting issues~\cite{murphy2012machine}. \\

In this paper, we consider a gray-box identification framework, relying on a port-Hamiltonian system (PHS) state-space representation and neural networks, known as port-Hamiltonian neural networks (PHNN) to identify the dynamics of a plucked nonlinear guitar string. PHS~\cite{maschke1992,van2014port} generalizes Hamiltonian mechanics to multi-physics open systems by explicitly modeling energy exchange with the 
environment through ports (inputs/outputs) and dissipation. While the PHS approach can be applied to simulate a wide range of physical domains, including acoustics, fluid mechanics, quantum physics and others~\cite{falaize2016passive,aoues2017modeling,cardoso2024port,roze2024time}, nonlinear PHS identification is still little explored (see e.g. Cherifi et al.~\cite{cherifi2020overview} for an overview),  and only a few examples applied to audio exist in the literature~\cite{najnudel2021identification}. PHNN leverage the PHS formulation and approximate each component of the system with multilayer perceptrons (MLPs), guaranteeing physical consistency and structure at both training and inference stages~\cite{desai2021, cherifi2025nonlinearporthamiltonianidentificationinputstateoutput,roth2025stableporthamiltonianneuralnetworks,moradi2026port,el2026ph}. Although some works use PHNN to model the well-known acoustic phenomena of self-oscillation~\cite{linares2026controlled}, the use of this framework to identify distributed audio systems, to the best of our knowledge, has not yet been described in the literature, where existing works are based on neural operators~\cite{schlecht2022physical,luan2025physics}, Koopman theory~\cite{diaz2024towards}, differentiable digital signal processing~\cite{lee2024differentiable} or modal synthesis methods~\cite{zheleznov2025learning,zheleznov2026stable,diaz2025fast}. \\

This work follows the ideas from Eidnes et al.~\cite{eidnes2024pseudo}, where they presented a general framework to learn Hamiltonian partial differential equations as finite-difference discretizations and is closely related to Zhelenov et al.~\cite{zheleznov2025learning,zheleznov2026stable}, where they learn the nonlinear string dynamics through a differentiable modal synthesis method. Lastly, we consider a time discretization method which provides an efficient, explicit, passivity-preserving solver by combining the scalar auxiliary variable (SAV) method~\cite{shen2018sav}, a non-iterative solver\cite{BILBAO2023111697}, and a drift-rejection control\cite{risse:hal-05222856}.  
This paper is organized as follows. Section \ref{sec: nonlinear string model} introduces the nonlinear string model under a continuous and semi-discrete port-Hamiltonian formulation, Section \ref{sec: string-phnn-model} presents the PHNN model used to identify the nonlinear string, Section \ref{sec: numerical-experiments} describes the numerical experiments, Section \ref{sec: results} discusses the results, and Section \ref{sec: conclusions} concludes the paper.
\section{Nonlinear 
string model}
\label{sec: nonlinear string model}

Consider the dynamics of a string under geometrically nonlinear conditions. The transverse displacement $q(x,t)$ of a string subject to large deformations is governed by the following PDE 
\begin{equation}
    \label{eq: string-pde}
    \partial_t^2q=\frac{1}{\mu}[(T\partial_x^2-EI\partial_x^4)q-2\mu(\eta_0-\eta_1\partial_x^2)\partial_tq+\frac{EA-T}{2}\partial_x(\partial_x q)^3+\delta(x-x_{e})f_{e}],
\end{equation}
where $q\in[0,l_0]\in\mathbb{R}$, for some string length $l_0$ [m], and $t\in\mathbb{R}^{+}$. The various constant parameters are the string tension $T$ [N], Young's modulus $E$ [Pa], radius $R$ [m], cross section $A=\pi R^2$ [$\text{m}^2$], moment of inertia $I=\pi R^4/4$ [$\text{m}^4$], density $\rho$ [$\text{kg}\cdot\text{m}^{-3}$], linear mass density $\mu=\rho A$ [$\text{kg}\cdot\text{m}^{-1}$], and dissipation coefficients $\eta_0$ and $\eta_1$. The control term $f(x,t)=\delta(x-x_e)f_{e}(t)$ [N] is applied at position $x=x_{e}$, where $\delta$ denotes a Diract delta function and $f_e$ simulates a pluck of a string~\cite{bilbao2019large}: $f_e(t)=\frac{1}{2}f_{amp}\left[1-\cos\left(\frac{\pi t}{T_e}\right) \right],~t\in[0,T_e];~0,~\text{otherwise}$, with $f_{amp}~[N]$ the excitation amplitude and $T_e~[\text{sec}]$ the excitation duration. Simply-supported boundary conditions $q(0,t)=0$ and $\partial_x^2q(0,t)=0$ are employed at the domain endpoints $x=0$ and $x=l_0$. Zero initial conditions are assumed (so that $q(x,0)=\partial_tq(x,0)=0$ for $x\in[0,l_0])$. This simple model contains some key elements of musical string modeling (large deformations, frequency dependent losses). For a more complete overview concerning musical string models, see Bilbao et Ducceschi~\cite{bilbao2023models}. Table \ref{tab: continuous-semi-discrete-formulation-nonlinear-string} shows the nonlinear string model written in the continuous and semi-discrete port-Hamiltonian PDE form. The (semi-)discrete formulation is obtained using structure preserving finite differences, with $\mathbf{q}\equiv[q_1,\cdots,q_{N-1}]$ transverse displacement and $\mathbf{p}\equiv[p_1,\cdots,p_{N-1}]$ momenta at the $N-1$ nodes of a spatial grid, where $q(x_s)=q_s$, $p(x_s)=p_s$, and a grid spacing $h$ such that $l_0=Nh$. \\

\begin{table}[h!]
\centering
\scalebox{0.8}{\begin{tabular}{| p{2cm} | p{6.8cm} | p{6.8cm} |}
\hline
\textbf{Category} & \textbf{Continuous PH Model} & \textbf{Semi-Discrete PH Model} \\
\hline

State \newline variables 
& $q(x,t)$ (displacement),\newline $p(x,t)$ (momentum) 
& $\mathbf{q} = [q_1,\dots,q_{N-1}]$, \newline $\mathbf{p} = [p_1,\dots,p_{N-1}]$ \\
\hline
Domain 
& $x \in [0,l_0],\ t \ge 0$ 
& Grid: $x_s = sh,\ l_0 = Nh$ \\
\hline 
Dynamics 
& Second-order nonlinear PDE 
& First-order ODE system \\
\hline
Hamiltonian 
& $\mathcal{H} = \int_0^{l_0} (\frac{p^2}{2\mu} + \frac{T}{2}(\partial_x q)^2 + \frac{EI}{2}(\partial_x^2 q)^2 + \frac{EA-T}{8}(\partial_x q)^4) dx$ 
& $H = \frac{\|\mathbf{p}\|^2}{2\mu} + \frac{T}{2}\|\mathbf{D}^- \mathbf{q}\|^2 + \frac{EI}{2}\|\mathbf{D}^2 \mathbf{q}\|^2 + \frac{EA-T}{8}\|(\mathbf{D}^- \mathbf{q})^{\circ 2}\|^2$,~with~$\|\mathbf{v}\|=\sqrt{h\mathbf{v}^T\mathbf{v}}$ \\
\hline
PHS form 
& $\begin{bmatrix}\partial_t q \\ \partial_t p \end{bmatrix}
= (J - R)\begin{bmatrix}\delta_q \mathcal{H} \\ \delta_p \mathcal{H}\end{bmatrix}
+ \begin{bmatrix}0 \\ \delta_{x_e}\end{bmatrix} f_e$ 
& $\begin{bmatrix}\partial_t \mathbf{q} \\ \partial_t \mathbf{p}\end{bmatrix}
= (\mathbf{J} - \mathbf{R})
\begin{bmatrix}\mathbf{Kq} + \mathbf{f}_{nl}(\mathbf{q}) \\ \mathbf{M}^{-1}\mathbf{p}\end{bmatrix}
+ \begin{bmatrix}0 \\ \mathbf{G}_p\end{bmatrix} \mathbf{u}$ \\
\hline
Structure matrix 
& $J = \begin{bmatrix}0 & 1 \\ -1 & 0\end{bmatrix}$ 
& $\mathbf{J} = \begin{bmatrix}0 & \mathbf{J}_0^T \\ -\mathbf{J}_0 & 0\end{bmatrix}$, $\mathbf{J}_0 = \frac{\mathbb{I}}{h}$ \\
\hline 
Dissipation 
& $R = \begin{bmatrix}0 & 0 \\ 0 & 2\mu(\eta_0 - \eta_1 \partial_x^2)\end{bmatrix}$ 
& $\mathbf{R} = \begin{bmatrix}0 & 0 \\ 0 & \mathbf{R}_0\end{bmatrix}$, 
$\mathbf{R}_0 = -\frac{2\mu}{h}\eta_1 \mathbf{D}^2 + \frac{2\mu}{h}\eta_0 \mathbb{I}$ \\
\hline  
Mass \newline operator 
& $\mu$ 
& $\mathbf{M}^{-1} = \frac{h}{\mu}\mathbb{I}$ \\
\hline 
Linear \newline operator 
& $(T\partial_x^2 - EI\partial_x^4)q$ 
& $\mathbf{K} = h(-T\mathbb{I} + EI\mathbf{D}^2)\mathbf{D}^2$ \\
\hline 
Nonlinear force 
& $\frac{EA-T}{2}\partial_x (\partial_x q)^3$ 
& $\mathbf{f}_{nl}(\mathbf{q}) = -h\frac{EA-T}{2}\mathbf{D}^+(\mathbf{D}^- \mathbf{q})^{\circ 3}$ \\
\hline 
Input 
& $\delta(x-x_e) f_e(t)$ 
& $\mathbf{u} = f_{e}(t),\ \mathbf{G}_p$ distributes input to $x_e$ \\
\hline 
Difference operators 
& Continuous derivatives $\partial_x,\partial_x^2,\partial^4_x$
& $[\mathbf{D}^{-}\mathbf{q}]_i=\frac{q_{i}-q_{i-1}}{h}$, $\mathbf{D}^+ = -(\mathbf{D}^-)^T$, \newline $\mathbf{D}^2 = \mathbf{D}^+\mathbf{D}^-$ \\

\hline 
\end{tabular}}
\caption{Nonlinear String Model: Continuous vs Semi-Discrete Port-Hamiltonian (PH) Formulation}
\label{tab: continuous-semi-discrete-formulation-nonlinear-string}
\end{table}
\section{String-PHNN model}
\label{sec: string-phnn-model}
\begin{figure}[h!]
\begin{minipage}{0.45\textwidth}
    \includegraphics[width=\linewidth]{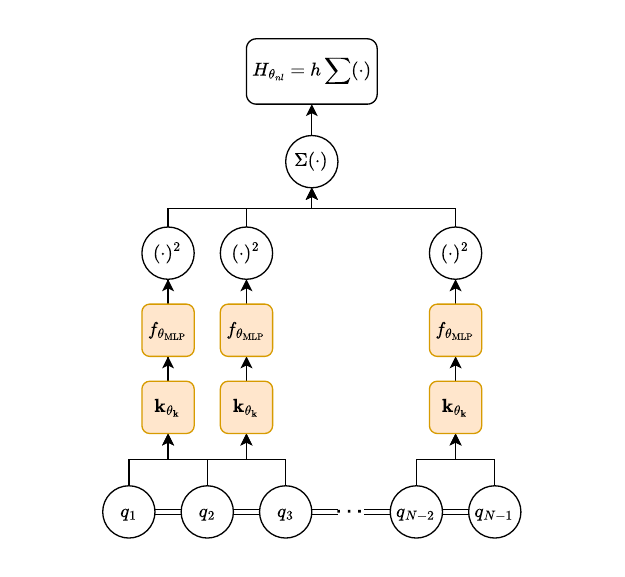}
    \caption{\centering The network $H_{\theta_{nl}}$ has one convolutional layer with kernel size two and a MLP with five hidden layers of 100 units and LeakyReLU activation.}
    \label{fig:phnn_string-hamiltonian}
\end{minipage}
\hspace{1cm}
\begin{minipage}{0.40\textwidth}
\centering
\centering
\scalebox{0.7}{\begin{tabular}{|c|c|c|c|c|}
\hline
                \multicolumn{5}{|c|}{\textbf{Nonlinear string configuration}} \\
                \hline
             \multicolumn{5}{|c|}{$\begin{gathered}
l_0=1.1~\text{m},\ \rho_0=8000~\text{kg}\cdot\text{m}^{-3},\ 
T=60~\text{N},\ E=2\times10^{11}~\text{Pa} \\
\eta_0=0.9~\text{s}^{-1},\ \eta_1=4\times10^{-4}~\text{m}^2\text{s}^{-1},\ N=202,\ h=5.4\times10^{-3}
\end{gathered}$} \\
            \hline
            \multirow{8}{*}{\rotatebox{90}{Dataset generation}} &
            \textbf{Parameter} & \textbf{Training} & \textbf{Validation} & \textbf{Test} \\
            \cline{2-5}
            & $N_{traj}$ & 48 & 12 & 60\\ 
            \cline{2-5}
            & $f_s$ & \multicolumn{3}{|c|}{88.2kHz} \\
            \cline{2-5}
            & $T_{s}$ & \multicolumn{3}{|c|}{2s} \\
            \cline{2-5}
            & $T_e$ & \multicolumn{3}{|c|}{[5,~30]~ms} \\
            \cline{2-5}
            & $x_{e}$ & \multicolumn{3}{|c|}{$\left[\frac{0.1l_0}{h},\frac{0.9l_0}{h}\right]$} \\
            \cline{2-5} 
            & $x_o$ & \multicolumn{3}{|c|}{$\lfloor\frac{\sqrt{2}l_0}{2h} \rfloor$} \\
            \cline{2-5} 
            &$f_{amp}$ & \multicolumn{3}{|c|}{[0.1,~5]~N}  \\
            \hline
            \multirow{4}{*}{\rotatebox{90}{ Hyperparameters}}
            & Batch size & 128 & - & - \\ 
            \cline{2-5} 
            & Learning rate & 1e-5 & - & - \\
            \cline{2-5} 
            & Optimizer [steps] & Adam [1M] & - & - \\
            \cline{2-5} 
            & Loss & \multicolumn{2}{|c|}{$\frac{1}{N} \sum_{i=1}^{N} \frac{\left| y_i - \hat{y}_i \right|}{dt}$} & $\frac{\sum_{i=1}^{N} (y_i - \hat{y}_i)^2}{\sum_{i=1}^{N} y_i^2}$ \\
            \hline
            \end{tabular}}
            \captionof{table}{Dataset generation parameters.}
            \label{tab:simulation_parameters}
\end{minipage}
\end{figure}

We parameterize the discrete Hamiltonian $H$ and the dissipation matrix $\mathbf{R}_0$ as 
\begin{equation}
    \label{eq: learnt_hamiltonian_pde}
    H_{\theta,\theta_{nl}}(\mathbf{q},\mathbf{p})=\frac{\|\mathbf{p}\|^2}{2\rho\pi R^2}+\frac{T}{2}\|\mathbf{D^{-}q}\|^2+\frac{E\pi R^4}{8}\|\mathbf{D}^2\mathbf{q} \|^2+H_{\theta_{nl}}(\mathbf{q}).
\end{equation}
and 
\begin{equation}
    \label{eq: learnt-dissipation-matrix}
    \mathbf{R}_{0,\theta}=-\frac{2\rho\pi R^2}{h}\eta_{1}\mathbf{D}^2+\frac{2\rho\pi R^2}{h}\eta_{0}\mathbb{I},
\end{equation}
where $\theta=(\rho,R,T,E,\eta_0,\eta_1)$ are positive learnable parameters and $H_{\theta_{nl}}(\mathbf{q}):\mathbb{R}^{N-1}\rightarrow\mathbb{R}$ is a neural network that approximates the non-quadratic part of the Hamiltonian. Note that learning $H_{\theta_{nl}}(\mathbf{q})$ poses the problem of how to approximate spatial derivatives by neural networks. In order to solve this, we follow the existing work in literature~\cite{celledoni2025predictions} where they highlight the connection between finite difference schemes and convolutional neural networks. For example, the first order backward finite difference operator
\begin{equation}
    [\mathbf{D}^{-}\mathbf{q}]_s=\frac{q_{s}-q_{s-1}}{h}=(\mathbf{k}*\mathbf{q})_s,
\end{equation}
where $\mathbf{k}=\frac{1}{h}[1,-1]$ is a kernel of size 2. In our case, it is natural to introduce this learning method to approximate the non-quadratic part of our target Hamiltonian $H_{nl}(\mathbf{q})$ as it can be written as 
\begin{equation}
    \label{eq: rewriting-hamiltonian}
    H_{nl}(\textbf{q})=\frac{EA-T}{8}\|(\textbf{D}^{-}\textbf{q})^{\circ 2} \|^2=h\sum_{s=1}^{N-1}\frac{EA-T}{8}\left(\frac{q_s-q_{s-1}}{h}\right)^4.
\end{equation}
Thus, we structure the network $H_{\theta_{nl}}(\mathbf{q})$ such that, for a given fixed $h$,
\begin{equation}
    H_{\theta_{nl}}(\mathbf{q})=h\sum_{s=1}^{N-1}\left(f_{\theta_{\text{MLP}}}(\mathbf{k}_{\theta_k}*\mathbf{q})\right)^2,
\end{equation}
which takes inputs of dimension $N-1$, the number of spatial discretizations points, and consists of one convolutional layer $\mathbf{k}_{\theta_k}=[k_{\theta_{k,1}},k_{\theta_{k,2}}]$ with kernel size two followed by an MLP $f_{\theta_{\text{MLP}}}$ identically applied to each of the outputs of the first convolutional layer. Finally, the last
layer performs a summation of the $N-1$ inputs to one scalar (see Figure \ref{fig:phnn_string-hamiltonian}). The structure of $H_{\theta_{nl}}(\mathbf{q})$, inspired from the work done in Eidnes et al.~\cite{eidnes2024pseudo}, is physically motivated by the structure of Hamiltonian PDEs: the spatial discretization of a Hamiltonian functional gives a Riemann sum over the spatial grid, the application of the homogeneity assumption allows the use of a shared convolutional operator combined with an MLP to model the local nonlinear energy density, and squaring the output guarantees non-negative local energy contributions. In the following, we refer to the combination of $H_{\theta,\theta_{nl}}$ and $\theta$ as StringPHNN.

\section{Numerical experiments}
\label{sec: numerical-experiments}

\subsection{Problem statement}
Given a discrete reference trajectory $\mathcal{T}$, the objective is to design a PHNN model capable of accurately generating a trajectory $\tilde{\mathcal{T}}$ that approximates $\mathcal{T}$ as closely as possible. The use of the modified SAV method requires to discretize the trajectories in time using staggered grids~\cite{Hairer2006} so that $\mathcal{T}=\{\mathbf{q}_s^{t-1/2},\mathbf{p}_s^t,f_e^{t+1/2},x_e \}$, for $s\in\{1,\cdots,N-1\}$ and $t\in\{0,\cdots,T_s/dt\}$. 

\begin{figure}[h!]
    \centering
    \includegraphics[width=0.7\linewidth]{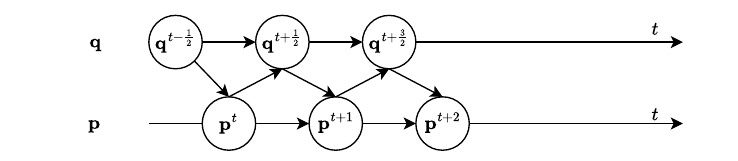}
    \caption{Representation of a staggered-in-time scheme, with variables defined at offset time instants.}
    \label{fig:placeholder}
\end{figure}

\subsection{Implementation details}

\subsubsection{Data generation}
We consider a nonlinear string with the fixed configuration detailed in Table \ref{tab:simulation_parameters}, which guarantees numerical stability for our choice of sampling frequency $f_s=88.2\text{kHz}$. A more detailed explanation on how to choose these parameters is given in Risse et al.~\cite{risse:hal-05222856}. The data generation procedure is similar to the one detailed in Zhelenov et al.~\cite{zheleznov2025learning} The simulation parameters used for the generation of the training, validation and testing datasets are outlined in Table \ref{tab:simulation_parameters}, where $N_{traj}$ is the number of trajectories and $T_{s}$ the duration of the simulation. Each trajectory consists of both the displacement $q_s$ and momentum $p_s$ information for each node $s$. The physical parameters correspond to a natural frequency $f_0\approx 55.5\text{Hz}$. This string is excited by randomly-generated excitation functions $f_e(t)$ at randomised excitation positions $x_e$ using a uniform distribution for the specified parameter ranges.

\subsubsection{Training and test details}
\begin{figure}[h!]
    \centering

    \begin{minipage}{0.49\textwidth}
        \includegraphics[width=\linewidth]{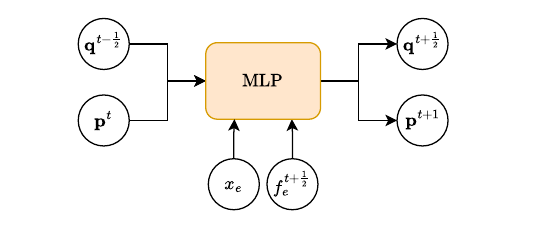}
    \caption{Computation graph of the baseline}
    \label{fig:computation_graph_baseline}
    \end{minipage}
    \hfill
    \begin{minipage}{0.49\textwidth}
        \includegraphics[width=\linewidth]{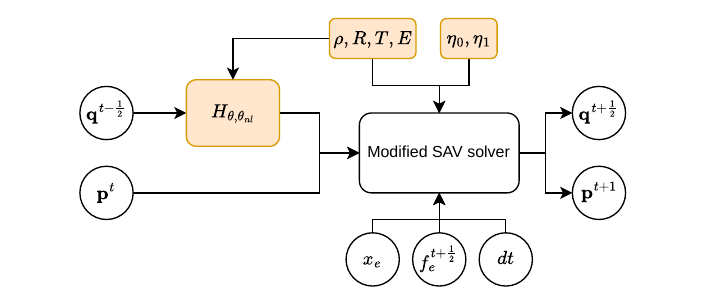}
    \caption{Computation graph of the StringPHNN}
    \label{fig:computation_graph_phnn}
    \end{minipage}
    
\end{figure}

The \textit{baseline} model consists of a MLP whose forward method directly approximates the discrete flow at the next time step $t+1$ (see Figure \ref{fig:computation_graph_baseline}), whereas the StringPHNN model implements the discrete PHS formulation in Table \ref{tab: continuous-semi-discrete-formulation-nonlinear-string}, with the network $H_{\theta_{nl}}$, learnable parameters $\theta=(\rho,R,T,E,\eta_0,\eta_1)$ and performs one time discretization step using the modified SAV method~\cite{risse:hal-05222856} (see Figure \ref{fig:computation_graph_phnn}). In both cases, the model takes as an input $\mathbf{y}^t=(\mathbf{q}^{t-1/2},\mathbf{p}^t),~f^{t+1/2},~x_e$ and returns as an output $\mathbf{\tilde{y}}^{t+1}=(\mathbf{\tilde{q}}^{t+1/2},\mathbf{\tilde{p}}^{t+1})$.  The training loss $\mathcal{L}_{\text{train}}(\mathbf{y}^{t+1},\mathbf{\tilde{y}}^{t+1})$ is the mean absolute error (MAE) divided by the time step $dt$ and, in the case of the StringPHNN, backpropagation is performed through the internal operations of the numerical method. We train all models using the Adam optimizer for $10^6$ optimization steps and repeat each experiment with 5 independent model initializations. All the experiments are carried out in the PyTorch framework\footnote{The code is publicly available in the GitHub repository: \hyperref[]{https://github.com/mlinaresv/LearningTheString}}. During testing, the models recursively generate the full trajectory from the 60 initial conditions provided in the dataset. The performance is evaluated using the relative mean squared error (MSE) $\mathcal{L}_{test}(\mathbf{y}^{t+1},\mathbf{\tilde{y}}^{t+1})$ computed over the entire trajectory. 

\section{Results and discussion}
\label{sec: results}
\begin{figure}[h!]
    \centering
    \begin{minipage}{0.43\textwidth}
        \centering
        \vspace{-0.40cm}
        \includegraphics[width=\linewidth]{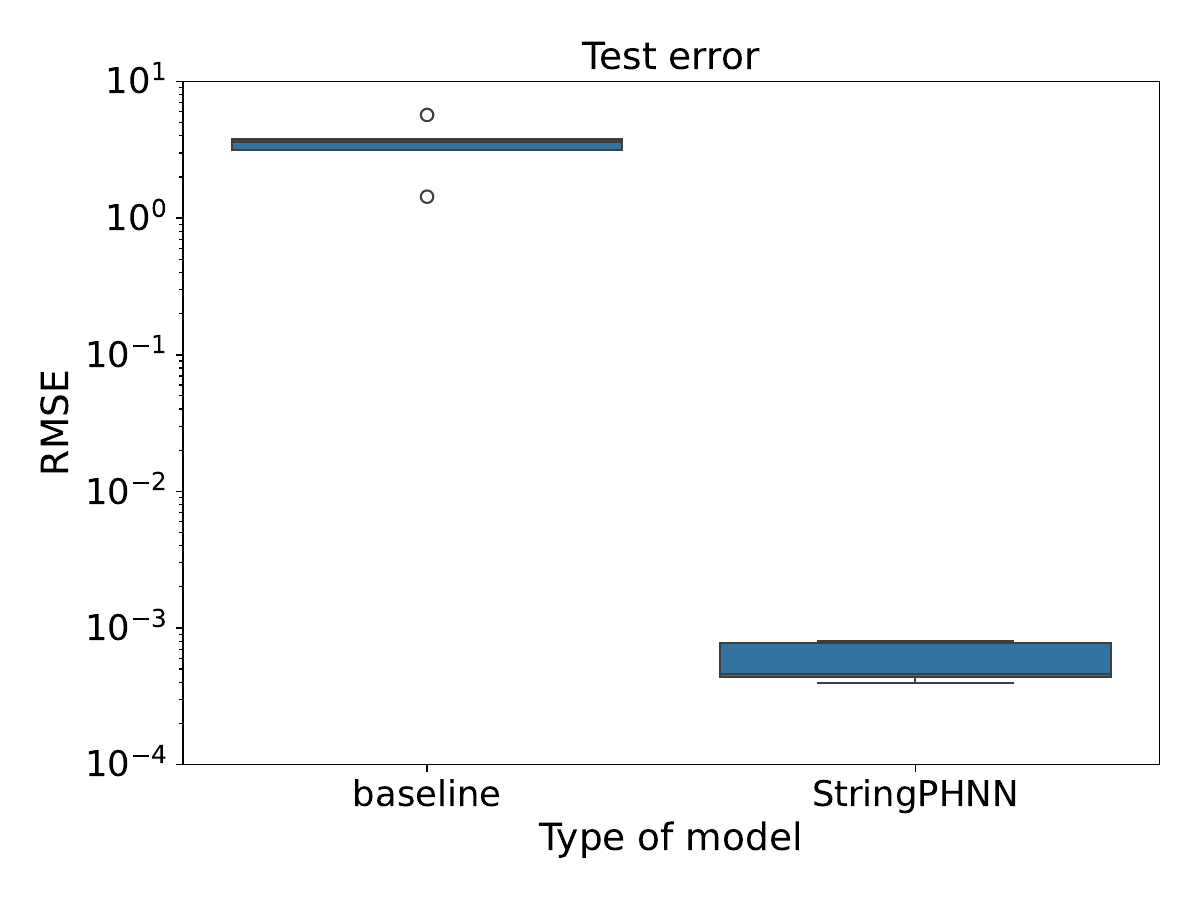}
        \caption{\centering Relative mean squared error (RMSE) of the baseline and StringPHNN model over the test dataset.}
        \label{fig:rmse_baseline_phnn}
    \end{minipage}
    \hfill
    \begin{minipage}{0.52\textwidth}
        
        \centering
        \includegraphics[width=\linewidth]{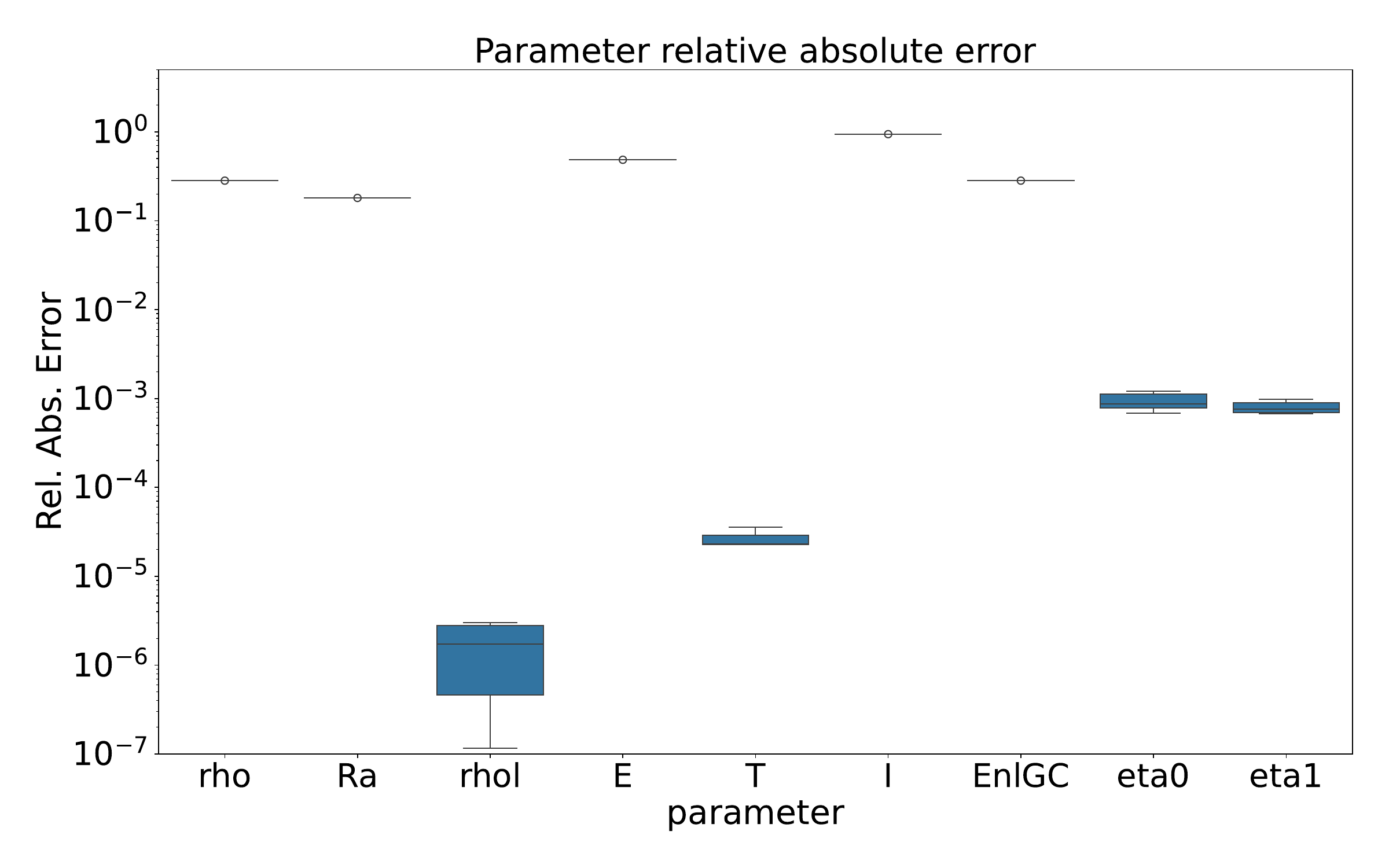}
        \caption{\centering Relative absolute error of the parameters learned by the StringPHNN. Legend: $\{\text{rho}:\rho,~\text{Ra}:R,~\text{rhol}:\pi R^2,~\text{E}:E,~\text{T}:T,~\text{I}:\pi R^4/4,~\text{EnlGC}=(EA-T)/8\},~\text{eta0}:\eta_0,~\text{eta1}:\eta_1\}$}
        \label{fig:relative-error-parameters}
    \end{minipage}
\end{figure}

Figure \ref{fig:rmse_baseline_phnn} compares the test RMSE obtained by the baseline model and the proposed StringPHNN. The baseline exhibits errors on the order of $10^0$, whereas the StringPHNN achieves errors around $10^{-4}$, outperforming the baseline by several orders of magnitude. Figure \ref{fig:relative-error-parameters} reports relative absolute error associated with the physical parameters identified by the StringPHNN. The results show that several parameters, such as the linear density $\mu$, the tension $T$, or the damping coefficients $\eta_0,\eta_1$ are recovered with very high accuracy. However, the mode struggles to identify parameters such as $\rho,R$ or $E$. These differences suggest that the way these physical quantities appeared in \eqref{eq: learnt_hamiltonian_pde} affects their identifiability from the observed dynamics. For example, $\mu=\rho\pi R^2$ is very well estimated even though $\rho$ and $R$ are not. A study on the non-dimensionalization rescaling could be carried out to mitigate this lack of uniqueness in terms of dynamical representation.

\begin{figure}[h!]
    \centering

        \centering
        \includegraphics[width=0.9\linewidth]{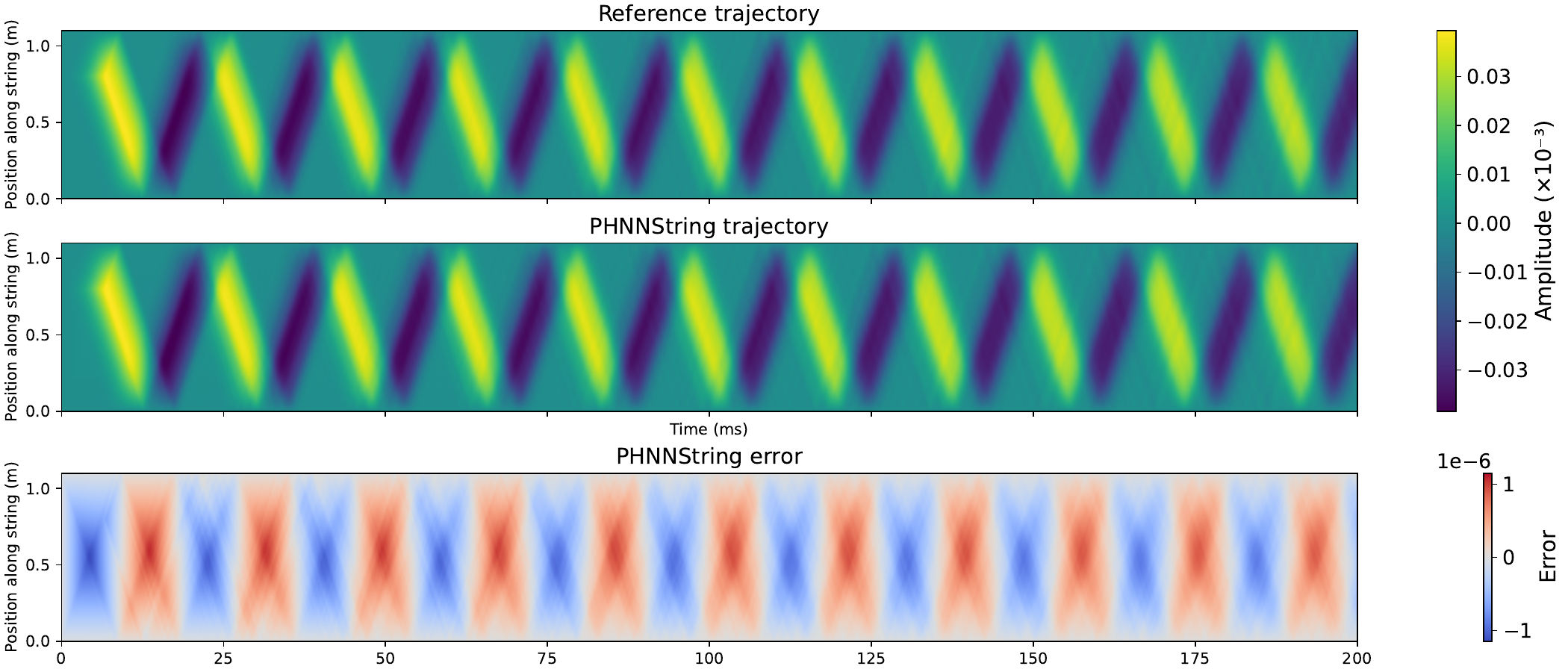}
        \caption{\centering  A reference displacement test trajectory compared with the PHNNString prediction (the initialization with the lowest test error is considered).}
        \label{fig: amplitude-error}
\end{figure}

\begin{figure}[h!]
        \centering

        \begin{subfigure}{0.32\textwidth}
            \centering
            \includegraphics[width=\linewidth]{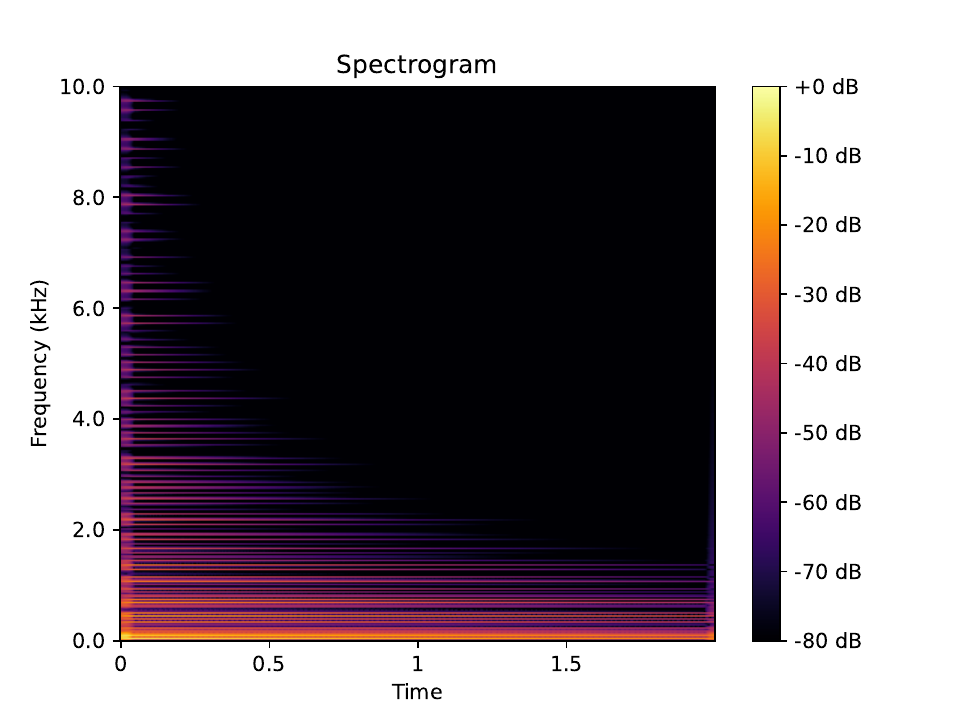}
            \caption{Reference spectrogram}
        \end{subfigure}
        \begin{subfigure}{0.32\textwidth}
            \centering
            \includegraphics[width=\linewidth]{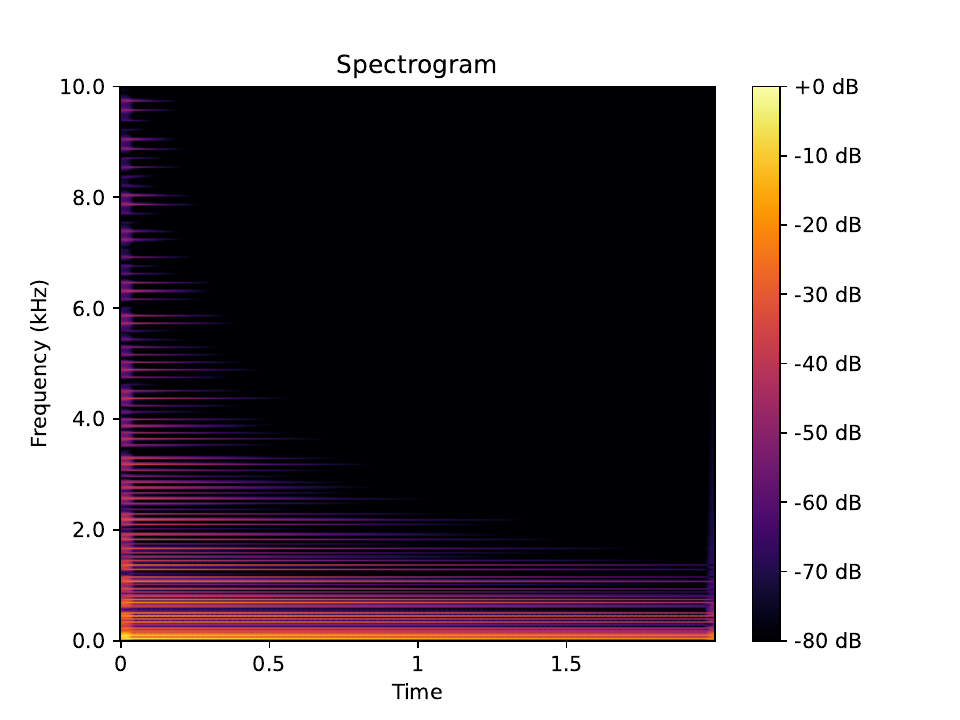}
            \caption{Predicted spectrogram}
        \end{subfigure}
        \begin{subfigure}{0.32\textwidth}
            \centering
            \includegraphics[width=\linewidth]{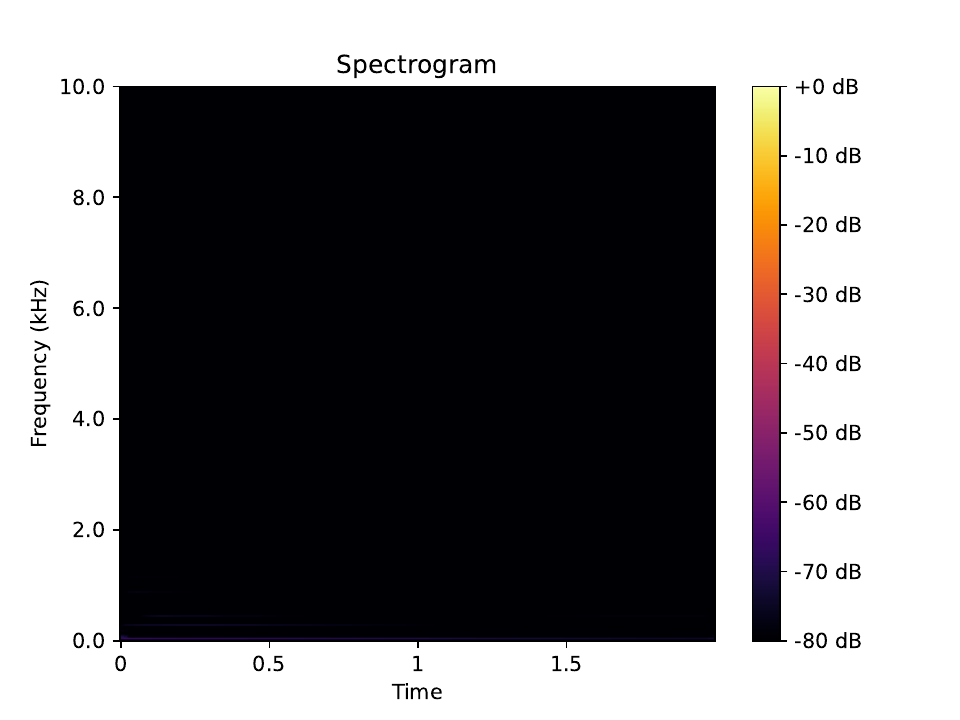}
            \caption{Error spectrogram}
        \end{subfigure}
        
        \caption{ \centering Spectrogram of a momentum test trajectory at the position $x_0$ (the initialization with the lowest test error is considered).}
        \label{fig: spectrogram}
\end{figure}

The ability of the model to capture the spatio-temporal propagation patterns is shown in Figure \ref{fig: amplitude-error}, where a reference displacement trajectory from the test dataset is compared with the prediction obtained using the PHNNString model. The predicted trajectory captures the wave dynamics and amplitude evolution along the string, with an error distribution remaining under $10^{-6}$. This is also reflected in terms of the spectral content as shown in Figure \ref{fig: spectrogram}, which presents a comparison between the reference and predicted spectrograms for a momentum test trajectory evaluated at the position $x_0$,

\section{Conclusion}
\label{sec: conclusions}
In this work, we introduced a neural network identification framework based on a energy-conserving numerical method for a port-Hamiltonian finite-difference discretization of a nonlinear string and compared it with a non-physics-informed baseline. Through our experiments, we showed that our StringPHNN outperformed the non-physics-informed baseline from the perspectives of accuracy and interpretability, as we do not only recover the reference dynamics but also its different components. Future research directions involve the identification of the nonlinear string dynamics from partial observations or measurements as well as from real audio recordings.

\section*{Acknowledgments} 

This project is co-funded by the European Union's Horizon Europe research and innovation program Cofund SOUND.AI under the Marie Sklodowska-Curie Grant Agreement No 101081674, and provided with computing HPC and storage resources by GENCI at IDRIS thanks to the grant 2026-105408 on the supercomputer Jean Zay's V100 partition. The authors are grateful to Sølve Eidnes and Benjamin Tapley for early insightful discussions on the model architecture, to Victor Zheleznov and Stefan Bilbao for their guidance and experience on string synthesis; and to Thomas Risse, for his very helpful SAV implementation code.

\bibliographystyle{IEEEtran}
\bibliography{BiblioDLPHS}

\end{document}